# Multimodality Biomedical Image Registration using Free Point Transformer Networks

Zachary M. C. Baum, Yipeng Hu, Dean C. Barratt

Centre for Medical Image Computing and Wellcome/EPSRC Centre for Interventional and Surgical Sciences, University College London, London, UK
zachary.baum.19@ucl.ac.uk

**Abstract.** We describe a point-set registration algorithm based on a novel free point transformer (FPT) network, designed for points extracted from multimodal biomedical images for registration tasks, such as those frequently encountered in ultrasound-guided interventional procedures. FPT is constructed with a global feature extractor which accepts unordered source and target point-sets of variable size. The extracted features are conditioned by a shared multilayer perceptron point transformer module to predict a displacement vector for each source point, transforming it into the target space. The point transformer module assumes no vicinity or smoothness in predicting spatial transformation and, together with the global feature extractor, is trained in a data-driven fashion with an unsupervised loss function. In a multimodal registration task using prostate MR and sparsely acquired ultrasound images, FPT yields comparable or improved results over other rigid and non-rigid registration methods. This demonstrates the versatility of FPT to learn registration directly from real, clinical training data and to generalize to a challenging task, such as the interventional application presented.

## 1 Introduction

Ultrasound imaging (US) is a widely used intraoperatively where real-time imaging is required. Owing to the difficulties in obtaining good quality diagnostic imaging which are associated with US, methods for image fusion between US and a second, usually preoperative, imaging modality are widely incorporated into image-guided interventions [1]. One use of multimodality image fusion is to provide magnetic resonance/transrectal ultrasound (MR-TRUS) fusion imaging during targeted prostate gland biopsies. MR-TRUS fusion superimposes the diagnostic information of magnetic resonance (MR) imaging on the transrectal ultrasound (TRUS) images. This enables clinicians to acquire samples from predefined lesions within the prostate in MR imaging and provides a real-time and low-cost solution that outperforms the current reference standard of US-guided systematic biopsy [2]. Furthermore, MR-TRUS fusion has been shown to improve the detection of high-grade prostate cancers and reduce sampling errors [3, 4]. Improved sampling benefits patient management as higher-risk patients are more likely to be identified and offered appropriate treatment options [4].

With its growing clinical use [5], the registration of pre-operative MR imaging to intraoperative TRUS persists as an active area of research [6-9]. Canonically, methods for MR-TRUS fusion must overcome the non-linear intensity differences between imaging modalities. Such methods must also be generalizable as to effectively handle inter- and intra-patient variation. Complete 3D US acquisition is often needed to obtain a full field of view that contains the prostate gland for registration [6-9]. While 3D US requires the probe to be held in place manually or with an additional robotic/mechanical device, 3D-to-2D registration methods utilize inherently 2D US, without the additional hardware requirements of 3D US acquisition [10-11]. However, recent advances in automatic, well-validated, learning-based segmentation methods for MR [12] and TRUS [12-13] permit real-time delineation of anatomical surfaces. Such surfaces may provide simplified representations for efficient and, perhaps more importantly, robust multimodal image registration in place of purely image-based methods.

Point-set registration is a widely-used and well-defined registration technique where a rigid or non-rigid spatial transformation model is defined and, subsequently, the optimal transformation is determined by a set of parameters for that model. Existing point-set registration algorithms, such as Iterative Closest Point (ICP) [14] and Coherent Point Drift (CPD) [15], use iterative optimization processes to determine the transformation for a given pair of point-sets [14-18]. In practice, the iterative nature of such methods may hinder their use in real-time registration tasks, leaving them unable to effectively take full advantage of the inherently real-time nature of US. Given the abilities for efficient inference and modeling complex, non-linear transformations, learning-based point-set registration can support rapid registration updates on-the-fly with sparse data – a task previously considered infeasible during time-critical interventional procedures with iterative registration methods.

In this work, we present a novel deep neural network architecture for data-driven, non-rigid point-set registration. The proposed Free Point Transformer (FPT) is trained in an unsupervised manner and therefore does not require ground-truth deformation data, which can be infeasible to obtain in interventional applications. FPT learns non-rigid transformation between multimodal images without any prior constraints such as displacement coherence or deformation smoothness. FPT also generalizes accurately to sparse point sets sampled from previously unseen patient data.

We present a quantitative analysis of FPT's performance in the MR-to-TRUS point-set registration task and compare it to other rigid and non-rigid registration methods. This work demonstrates FPT's feasibility for continual real-time MR-TRUS fusion in prostate biopsy using sparse data which may be generated from automatically segmented sagittal and transverse slices available from existing bi-plane TRUS probes.

## 2    Methods

### 2.1    Network Architecture

FPT is composed of two modules, a global feature extractor, and a point transformer, as illustrated in Fig. 1.

The first module, the global feature extractor, accepts two point-sets: the target point-set, $P_T$, and the source point-set, $P_S$ and serves to extract permutation invariant and rotation invariant features from the point-sets. This module was composed of twin weight-sharing PointNets. PointNet is a previously-proposed neural network architecture that operates on a single point-set and allows permutation invariance [19], which has transformed how point-sets are represented and interpreted in many computer vision tasks, such as classification and segmentation. In this work, the 'input and feature transformation' and the 'global information aggregation' components of the original PointNet [19] are utilized as our global feature extractor. The global feature extractor module allows FPT to create a permutation and transformation invariant embedding function. Weights are shared between each PointNet to ensure that the inputs to the network pass through the same embedding function, which then aggregates each input into a 1024-dimensional source and target feature vector, respectively. PointNet's 'T-net' modules ensure FPT only learns transformations between point-sets which are relevant to the task, by applying a $3 \times 3$ transformation matrix to the coordinates of the input points [19]. The source and target feature vectors are concatenated into a 2048-dimensional global feature vector.

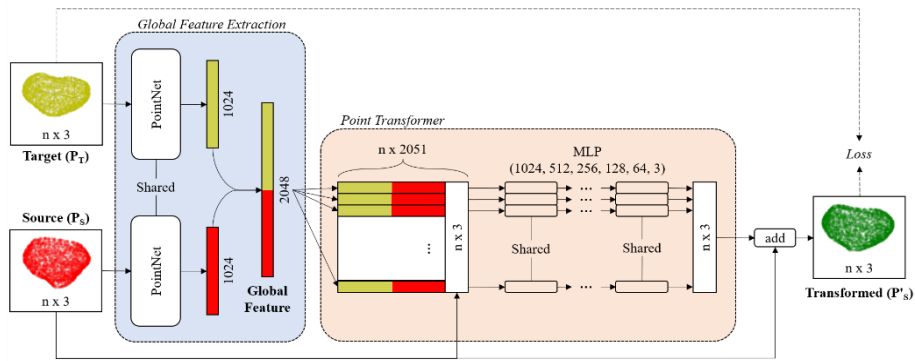

**Fig. 1.** Schematic representation of the FPT architecture for non-rigid point-set registration.

The second module, the point transformer, contains a series of weight-shared multilayer perceptrons (MLP). Each layer of the 6 layers consisted of a group of 2048 weight-shared fully connected layers with 1024, 512, 256, 128, 64, and 3 nodes per layer (Fig. 1). The first 5 layers used the ReLU activation function and the final layer used a linear activation function. This MLP is implemented as a series of 1D convolutions with a kernel size of one. This weight-sharing design choice allows potential regularization benefits for generalization, and ensures that each point passes through the point transformer via a common transformation function. The global feature vector is concatenated with each of the x, y, and z point location coordinates and passed through the MLP to produce an independent displacement vector for each point in $P_S$. This allows the global feature vector to be combined with all points in $P_S$, yet predict the displacements at individual locations independently using only the input point-sets. As such, the point transformer transforms each point without any constraints on smoothness or spatial coherence. The point transformer is only conditioned on the input feature vectors to determine a "model-free" transformation. Finally, the displacement vectors are added to $P_S$, to yield the transformed point-set, $P_S'$, upon which a loss may be computed.

### 2.2 Loss Function

Instead of an often-constrained spatial transformation model, FPT utilized a data-driven strategy to predict a displacement field on unstructured point locations. Prior knowledge regarding outliers, missing data and noise can be handled through data augmentation, reduction, and perturbation of the training data, respectively. Among distance metrics that do not require established correspondence, many require additional parameter tuning and explicit consideration of outliers or noise levels, such as those based on the likelihood or the divergence between point distributions.

In this work, we train using sparse data to illustrate the efficacy of our data-driven approach with a Chamfer distance [20]. We utilized the Chamfer distance as the basis for our loss function as it is simple to compute and easily parallelizable [20]. Our implementation has adapted the original Chamfer distance to a two-way formulation that minimizes mean distances between nearest neighbors in $P_T$ and $P_S'$. However, other types of metrics and possible loss functions warrant investigation in future studies.

### 2.3 Implementation Details

FPT was trained using the ModelNet40 [21] dataset was used to pre-train FPT. This was done with a minibatch size of 32 and a learning rate of $10^{-3}$ with the Adam optimizer. By pre-training with a large dataset, we leverage what was learned with ModelNet40 to improve generalizability in another setting [22]. ModelNet40 contains meshes of 40 distinct shapes which are randomly split into a 9843 model training set and a 2468 model testing set. The point-sets are a collection of 2048 points uniformly sampled from these mesh surfaces. In training, the point-sets were augmented on-the-fly with scaling, deformation, and a transformation comprised of rotation and displacement. Point-sets were scaled, per-sample, between [-1, 1]. The scaled input is used as $P_T$. We simulated the non-rigid transformations on the scaled point-sets by TPS transformation. TPS deformation was defined by a perturbation of the control points by Gaussian random shift. Rotation angles for the transformation were randomly sampled from [-45°, 45°] about each axis, with displacements randomly sampled from [-1, 1] in each of the X, Y, and Z directions. The scaled, deformed, and transformed version of the input was used as $P_S$. The known transformations were only used for validation, as training was unsupervised.

## 3 Experiments

### 3.1 Data

The experimental dataset used in our evaluation was comprised of 108 pairs of pre-operative T2-weighted MR and intraoperative TRUS images from 76 patients which were acquired during the Smart Target clinical trials [23]. The dataset was split into training and testing sets, each containing 54 (50%) of the 108 patient pairs. Given its data-driven architecture, FPT was not defined by any hyperparameters beyond those described in Section 2.3. In this work, we did not use a hold-out set to prevent bias through an exhaustive hyperparameter search when fine-tuning the networks for the experiments described in Section 3.3. Therefore, this two-way random split experiment provided a non-overfitted estimate of registration performance, although data from different centers or differing acquisition protocols are still of value for future validation.

### 3.2 Implementation Details

In each experiment, the performance on the MR to TRUS registration task was evaluated using four different methods: center-alignment, ICP, CPD, and FPT. Center-alignment simply involved aligning the mean of each input point-set at the origin. ICP [14] is a widely-used, iterative method for rigid point-set registration. CPD [15] is a widely-validated, non-rigid, and iterative point-set registration algorithm.

As we sought to demonstrate the feasibility of FPT, we did not perform an exhaustive search of hyperparameter combinations for all methods to which ours is compared. All settings and implementation details that provided the best results in our search for each method are reported below. ICP was allowed to run for up to 25 iterations, all other parameters or initializations were performed as described in [14]. CPD was performed with $w = 0$, making the weight of the uniform distribution zero. We permitted CPD to run for up to 150 iterations. All other parameters remained as default [15]. The use of potentially non-optimized ICP and CPD also demonstrates the importance of initialization and parameter-tuning for such methods.

FPT was tested in two variations. First, where the network was only pre-trained on ModelNet40, as described in Section 2.3, and second, where the ModelNet40-trained network was fine-tuned on the MR-TRUS training dataset. Fine-tuning was performed with the same parameters which were used in pre-training. When fine-tuning, $P_T$ was defined as the normalized TRUS prostate surface points, and $P_S$ was defined as the normalized MR prostate surface points. No deformation, translation, or rotation was added to the surface points. No trainable weights of the networks were frozen.

### 3.3 Experimental Protocol

**MR to TRUS Registration**
In the first experiment, we presented each method with all TRUS and MR surface points to assess each method with complete data. For FPT, fine-tuning was performed with the training set. All evaluations were performed using only the testing set.

**MR to Sparse TRUS Registration**
In the second experiment, we assessed the performance of each method using sparse TRUS surface points to reflect a plausible clinical scenario, as described in Section 1. To simulate sparse TRUS data, we simulated TRUS surface points captured from one simultaneous acquisition from a biplane TRUS in the sagittal and transverse planes (Fig. 2). This was done by removing the TRUS surface points which would not be visible in one simultaneous acquisition. As with our first experiment, fine-tuning was performed with the training set, and all evaluations were performed with the testing set.

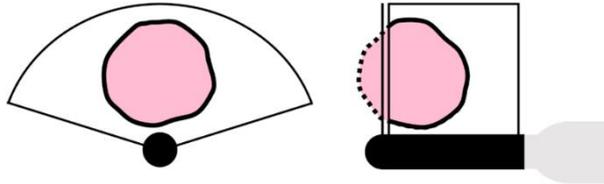

**Fig. 2.** Illustration of contours from which surface points would be extracted from a biplane TRUS transducer. Points from the transverse plane (left) and sagittal plane (right) that would be used are shown with solid lines. Dashed lines and other surface points are discarded.

**Evaluation Metrics**

All registrations were evaluated on their displacement predictions using Chamfer distance ($D_C$), Hausdorff distance ($D_H$), and registration time. We report registration accuracy on independent landmarks with target registration error (TRE) as has been used in many prior studies validation multimodal image registration [6-9], where its clinical relevance has been established. TRE is defined as the root-mean-square of the distances computed between all pairs of registered source and target landmarks for each patient. The landmarks comprised of 145 pairs of points included the apex and base of the prostate and patient-specific landmarks such as zonal structure boundaries, water-filled cysts, and calcifications, the spatial distribution of which is representative of the target registration distribution in this application. Landmarks were not included in any training, fine-tuning, or registration processes.

## 4 Results

Our quantitative results for the first experiment (Table 1) demonstrate a fine-tuned FPT's comparable or improved results to ICP and CPD in all metrics. Example qualitative results from the first experiment are provided in Figure 3.

**Table 1.** Results from the first experiment, using complete TRUS data. STD: Standard Deviation.

| Methods | Time (s) Mean | $D_C$ (mm) Mean ± STD | $D_H$ (mm) Mean ± STD | TRE (mm) Mean ± STD |
|---|---|---|---|---|
| Center-aligned | - | 2.7 ± 0.9 | 10.7 ± 2.6 | 5.1 ± 1.7 |
| ICP [14] | 0.15 | 2.5 ± 1.0 | 10.4 ± 2.5 | 4.9 ± 1.8 |
| CPD [15] | 13.77 | **1.2 ± 0.2** | 6.5 ± 1.6 | 4.8 ± 1.9 |
| FPT | **0.08** | 1.7 ± 0.3 | 8.2 ± 2.1 | 4.9 ± 1.9 |
| FPT (Fine Tuned) | **0.08** | 1.5 ± 0.2 | **6.3 ± 1.6** | **4.7 ± 1.8** |

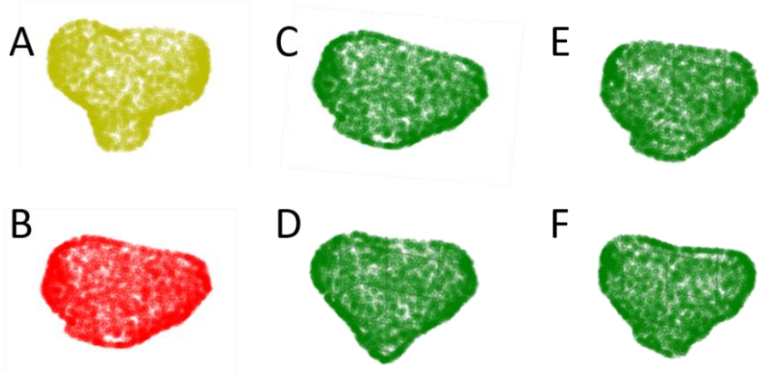

**Fig. 3.** Registration results from the first experiment on a target (a) and source (b) point-set using ICP (c), CPD (d), FPT (e), and FPT after fine-tuning (f).

In the second experiment (Table 2), a fine-tuned FPT demonstrates improved results across all metrics, compared with those from ICP and CPD. Fine-tuning also further improves the registrations with respect to $D_C$ and $D_H$. Comparing inference times, FPT requires, on average, 0.08 s per registration, compared with 0.14 s and 11 s, for ICP and CPD, respectively. Example qualitative results from the second experiment are provided in Figure 4.

**Table 2.** Results from the second experiment, using sparse TRUS data. STD: Standard Deviation.

| Methods | Time (s) Mean | $D_C$ (mm) Mean ± STD | $D_H$ (mm) Mean ± STD | TRE (mm) Mean ± STD |
|---|---|---|---|---|
| Center-aligned | - | 2.9 ± 1.0 | 10.9 ± 2.7 | 5.3 ± 1.7 |
| ICP [14] | 0.13 | 2.6 ± 1.0 | 10.5 ± 2.4 | 5.0 ± 1.7 |
| CPD [15] | 11.08 | 7.2 ± 1.4 | 19.8 ± 4.5 | 6.9 ± 2.7 |
| FPT | **0.08** | 4.1 ± 0.8 | 11.3 ± 2.8 | **4.4 ± 1.4** |
| FPT (Fine Tuned) | **0.08** | **1.9 ± 0.3** | **6.8 ± 1.4** | 4.9 ± 1.7 |

## 5   Discussion

Conventional intensity-based registration algorithms for MR-TRUS fusion samples intensity information directly, whereas FPT receives only geometric and spatial information from the surface point-sets in the form of very limited and, potentially, easy-to-acquire data 2D US slices. Recently, conventional methods have obtained TREs of 1.5 mm [6], 2.4 mm [7], 1.9 mm [8], or 3.6 mm [9], validated on 16, 8, 8, and 76 patients respectively. Potentially, the robustness of point-set extraction from prostate gland segmentation may reduce variance in registration error, although additional validation is needed to draw further conclusions. However, comparing to other iterative or learning-based intensity-based registration methods may be considered outside of the scope of this work, due to the specific clinical scenarios of interest such as sparse slice availability. Nonetheless, our results demonstrate that FPT can directly learn descriptive and data-driven features from sparse data. From these features, FPT can efficiently compute a set of accurate displacements, as compared to conventional image-based registration methods. Further validation and investigation are required to assess FPT's ability to generalize on multi-center data, wherein there may be increased data heterogeneity.

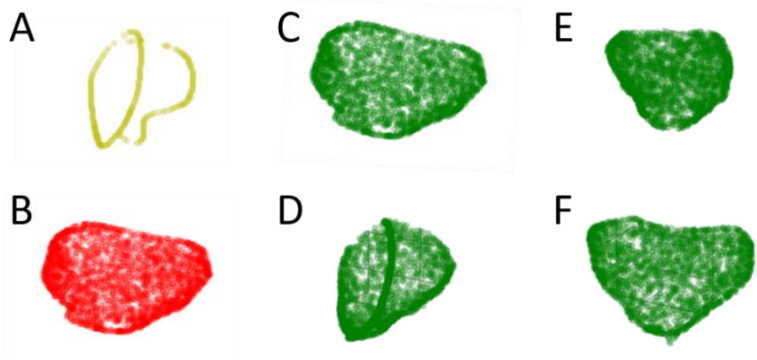

**Fig. 4.** Registration results from the second experiment on a sparse target (a) and source (b) point-set using ICP (c), CPD (d), FPT (e), and FPT after fine-tuning (f). The complete surface of (a) is identical to Figure 3a.

Without fine-tuning, we see that FPT demonstrates the lowest TRE on the test dataset. It is possible that fine-tuning may result in FPT overfitting the training dataset, yielding a higher TRE. However, the fine-tuned FPT greatly outperforms the non-fine-tuned FPT in $D_C$, upon which it is trained to minimize, and $D_H$.

Given its rapid point-set registration approach, FPT may serve other multimodality registration applications, such as computed tomography/US (CT-US) fusion, well. As previously described with MR-TRUS fusion, CT-US fusion is an active area of research; with its use ranging from surgical interventions [24-25] to radiotherapy planning [26]. As such, non-rigid point-set registration of surfaces extracted from US and CT may provide useful intraoperative visualizations which are of interest in future work, given the results in this work for prostate with MR-TRUS fusion.

## 6 Conclusion

We have presented Free Point Transformer (FPT), a deep neural network architecture for unsupervised data-driven point-set registration. FPT learns the displacement field required to produce individual point displacements using only the geometric information of its inputs. Evaluated on a real-world MR to TRUS registration task, FPT yields improvements or comparable performance to Iterative Closest Point and Coherent Point Drift. Most saliently, this work demonstrates that with a variable point-set sparsity, which may be generated from automatically segmented sagittal and transverse slices, readily available for all existing bi-plane TRUS probes in realistic clinical practices, FPT may enable continual real-time MR-TRUS fusion during prostate biopsies.

**Acknowledgments.** Z. Baum is supported by the Natural Sciences and Engineering Research Council of Canada Postgraduate Scholarships-Doctoral Program, the University College London Overseas and Graduate Research Scholarships. This work is also supported by the Wellcome/EPSRC Centre for Interventional and Surgical Sciences (203145Z/16/Z).